\documentclass{article}
\usepackage{spconf}
\usepackage{amsmath}
\usepackage{amsfonts}
\usepackage{graphicx}
\usepackage{booktabs}
\usepackage{url}
\usepackage[table]{xcolor}
\usepackage[margin=1in]{geometry}
\usepackage{tabularx}
\usepackage{enumitem}
\usepackage{ragged2e}
\usepackage{boldline}
\usepackage[T1]{fontenc}
\usepackage{float}
\title{Layer-wise Pruning of Transformer Attention Heads\\for Efficient Language Modeling}

\name{
Kyuhong Shim$^{1}$ \quad 
Iksoo Choi$^{1}$ \quad 
Wonyong Sung$^{1}$ \quad 
Jungwook Choi$^{2}$
}
  
\address{
$^{1}$Seoul National University \quad $^{2}$Hanyang University \\
\tt{\{skhu20, akacis, wysung\}@snu.ac.kr, choij@hanyang.ac.kr}
}

\begin{document}
%\ninept
\maketitle
% =========================================================================================== %
\begin{abstract}
While Transformer-based models have shown impressive language modeling performance, the large computation cost is often prohibitive for practical use. Attention head pruning, which removes unnecessary attention heads in the multihead attention, is a promising technique to solve this problem. However, it does not evenly reduce the overall load because the heavy feedforward module is not affected by head pruning. In this paper, we apply layer-wise attention head pruning on \emph{All-attention} \cite{sukhbaatar2019augmenting} Transformer so that the entire computation and the number of parameters can be reduced proportionally to the number of pruned heads. While the architecture has the potential to fully utilize head pruning, we propose three training methods that are especially helpful to minimize performance degradation and stabilize the pruning process. Our pruned model shows consistently lower perplexity within a comparable parameter size than Transformer-XL on WikiText-103 language modeling benchmark.

\end{abstract}
% =========================================================================================== %
\section{Introduction}
\label{sec:introduction}
Transformer-based neural networks \cite{vaswani2017attention} have been widely used for their great capability to capture long-range contextual relationships. Transformer models have achieved state-of-the-art performance in various tasks using sequential data, such as language modeling (LM) \cite{dai2019transformer, sukhbaatar2019adaptive} or language representation learning \cite{devlin2019bert, brown2020language}.

Transformer layer consists of two sub-modules: a multihead attention module (MHA) followed by a feedforward module (FF). Both components behave similarly in that they transform representations, but the information they use is dearly distinct. MHA extracts features based on the relationship between sequential inputs while FF transforms the feature irrespective of its relative location and value. In MHA, the connection between each input is measured from multiple perspectives by dividing the features into several attention heads. It has been reported that each head focuses on different parts of the sequence \cite{clark2019does}. Concurrently, it has been also presented that a considerable number of heads can be removed without performance loss \cite{voita2019analyzing, michel2019sixteen, hao2021self}.

\begin{figure}[t]
    \vspace{5pt}
    \centering
    \includegraphics[width=\linewidth]{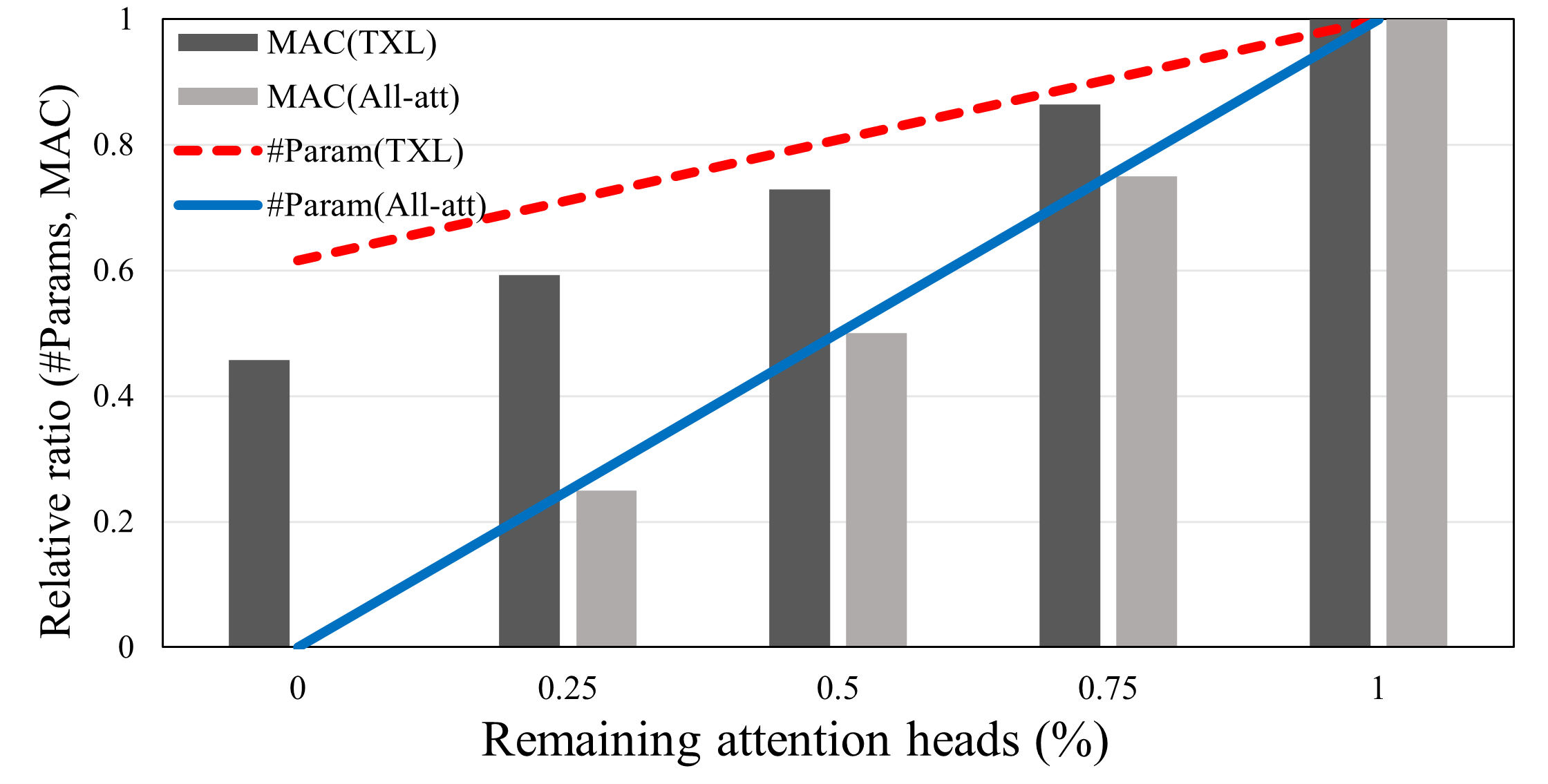}
    \caption{Relative computation and the number of parameters reduced by attention head pruning. Transformer-XL and All-attention Transformer are compared.}
    \label{fig:comparison}
    \vspace{-15pt}
\end{figure}

Despite its excellent performance, the computational cost and the parameter size of Transformer are considerably large. Attention head pruning is a promising method for reducing both. This is a structured pruning approach, so the effect of pruning can be well reflected in practical usage on modern devices, in contrast to unstructured pruning. However, the benefit only applies to MHA because FF is not affected by the number of heads, while FF often takes approximately 2/3 of the parameters and half of the computations (depending on the sequence length and model configuration). To further extend the ability to compress Transformer models with attention head pruning, we adopt the recently introduced \emph{All-attention} \cite{sukhbaatar2019augmenting} Transformer, which adds persistent memory blocks inside MHA, instead of FF. We denote All-attention Transformer as All-att for simplicity.

All-att unifies two sub-modules in the original Transformer and split almost every computation under the multihead path, which is a desirable characteristic for attention head pruning. Figure \ref{fig:comparison} demonstrates the advantage of the attention head pruning on All-att compared to a Transformer-XL (TXL) \cite{dai2019transformer}, which is a widely adopted model for LM. For example, in 50\% head sparsity, TXL computes approximately 73\% of full multiply-accumulate operations (MAC) and maintains 81\% of the parameters, whereas All-att only requires 50\% of the load and 50\% of the parameters.

In pruning attention heads, we utilize a trainable method so the model can jointly learn which heads can be pruned out while preserving the performance. Specifically, we attach auxiliary gating parameters on each layer, inspired by earlier works \cite{voita2019analyzing, bejnordi2019batch}. Although All-att shows comparable performance to the original Transformer in LM, removing each attention head of All-att is directly connected to losing the information inside persistent memory which replaces the role of the FF. We identify several difficulties in the pruning process; severe instability at the initial stage, consistent increase of the training loss, overly sparse heads, and significant performance drop of the pruned model. Therefore, we propose three techniques that modify the pruning process to solve these problems: (1) sparsity loss warm-up, (2) proper initialization, and (3) attention output scaling.

Our main contributions are summarized as follows. First, we adopt All-att to fully utilize the advantages of attention head pruning. Second, we propose advanced training techniques to minimize the damage to the performance of the pruned model and stabilize the pruning process. We demonstrate that our pruned All-att model shows consistently lower perplexity for word-level LM and lower bit-per-character for character-level LM, compared to the original Transformer model of a comparable parameter size.

% =========================================================================================== %
\section{Related Work}
\label{sec:related}
Pruning on Transformer has been widely studied. Research on unstructured pruning \cite{guo2020parameter, sanh2020movement} shows that several parameters can be removed without a significant effect on the final performance. However, the unstructured nature is practically difficult to take advantage of the actual speedup without specialized hardware support \cite{wang2021spatten}. 

Several studies have focused on attention head removal, which is a structured and GPU-friendly approach. The most adopted method begins from a fully converged pretrained model and prunes out attention heads during additional training steps. For example, in \cite{voita2019analyzing}, trainable gating parameters are attached to each head and regularized with $L_0$ loss. Other types of head pruning have also been proposed; without additional parameters, in \cite{michel2019sixteen}, the sensitivity of each head to the loss is used as a proxy for importance. A single-shot meta-pruner \cite{zhang2021know} is introduced in which a small convolutional neural network is trained to select heads that contribute to maintaining the attention distribution. Because earlier studies on attention head pruning do not compress FF, additional effort is needed to further reduce the computation and parameters of FF for the original Transformer.

% =========================================================================================== %
\section{Attention Head Pruning for LM}
\label{sec:method}
\begin{figure}[t]
  \centering
  \vspace{5pt}
  \includegraphics[width=0.9\linewidth]{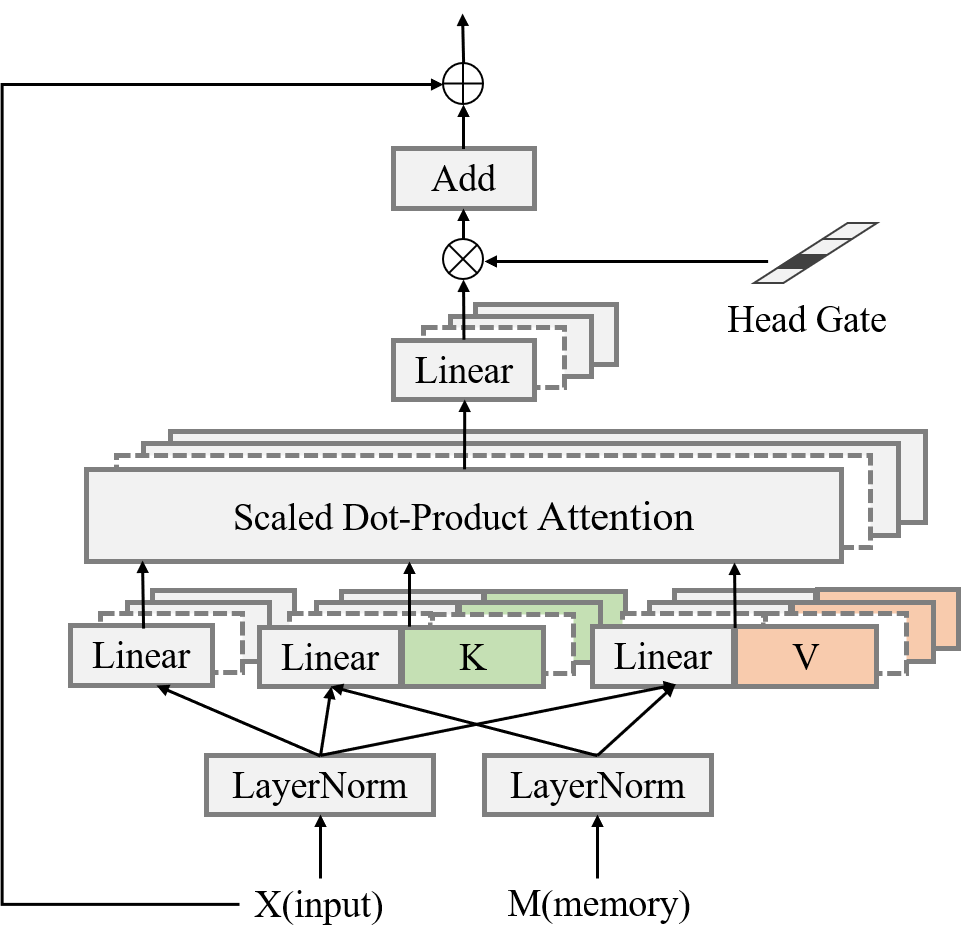}
  \caption{All-attention Transformer with a head gating module. K and V indicates persistent vectors that replace the role of the feedforward in original Transformer. The parts that disappear (dashed boxes) represent the saved operations of erased attention heads.}
  \label{fig:architecture}
  \vspace{-10pt}
\end{figure}

% ======================================================================================== %
\subsection{All-Attention Transformer}

All-att adds a set of trainable parameters, named as persistent vectors, instead of FF. These persistent vectors perform as an external key and value for Transformer but do not depend on the input. Figure \ref{fig:architecture} illustrates an All-att layer. For simplicity, we omit the relative positional encoding and its projection in equations and the figure.

When All-att architecture is used for LM, the memory caching algorithm of TXL is adopted. The hidden representation computed for the previous sequence segment is cached as memory and used as an external source of information. This memory mechanism enables much longer context, which is highly beneficial for LM. Consider a sequence of $d$-dimensional vector $X\text{=}\{x_1, x_2, ... x_T\}$ and a corresponding memory $M\text{=}\{m_1, m_2, ... m_S\}$. The query ($Q$), key ($K$), and value ($V$) of $i$-th head is calculated as $Q_i = \textbf{W}_i^Q X,\text{ } K_i = \textbf{W}_i^K [X, M],\text{ } V_i = \textbf{W}_i^V [X, M]$. The concatenation operator is noted as $[\text{ , }]$. For the entire model, $H$ heads are used per layer and ${L}$ layers are stacked.

The persistent vectors are realized as $N$ trainable $d_h$-dimensional vectors for each head, where $d_h\text{=}d/H$ is the head dimension. $\textbf{P}_i^K\text{=}\{p_{i,1}^k, ... p_{i,N}^k \}$ and $\textbf{P}_i^V\text{=}\{p_{i,1}^v, ... p_{i,N}^v\}$ represent the persistent key and value vectors of the $i$-th head. Every query in the sequence treats $\textbf{P}_i^K$ and $\textbf{P}_i^V$ as extensions of $K$ and $V$, respectively. The output of the $i$-th head is calculated as:

\vspace{-10pt}
\begin{equation}
    h_i = \text{Softmax}\left( {{Q_i}{[K_i, \textbf{P}_i^K]^T}} / {\sqrt{d_h}} \right)[V_i, \textbf{P}_i^V]
\end{equation}

The outputs from multiple attention heads are concatenated and projected to produce the final result.

\vspace{-10pt}
\begin{equation}
  O = \textbf{W}^O [h_1, h_2, ... h_H] = \sum_{i=1}^H \textbf{W}_i^O h_i
\end{equation}

By setting the number of persistent vectors $N$ same as the internal dimension of FF, the number of parameters of All-att becomes almost identical to that of the original Transformer (both MHA and FF).

% ======================================================================================== %

\subsection{Head Pruning by Gating}

For pruning, we attach a set of trainable head gating parameters $\Pi\text{=}\{\pi_1, \pi_2, .. \pi_H\}$ ($\pi_i \in \mathbb{R} $) to each layer. The parameters pass through \emph{BinConcrete} \cite{louizos2018learning} function and converted to stochastic discrete Bernoulli gate $G\text{=}\{g_1, g_2, ... g_H\}$ ($g_i \in \{0, 1\}$). The final projection is modified as follows:

\vspace{-10pt}
\begin{equation}
  O = s_g\sum_{i=1}^H g_i h_i\textbf{W}_i^O = \frac{H}{\sum_{i=1}^H g_i}\sum_{i=1}^H g_i h_i\textbf{W}_i^O
  \label{equation:gating_projection}
\end{equation}

To avoid division by zero (if all gates are sampled to 0), we clip $s_g$ to the maximum value of $H$. Because we can easily absorb $s_g$ in the $W^O$, the scaling $s_g$ does not require additional computation for the inference.

In addition to the default negative log-likelihood (\emph{nll}) loss, we utilize additional $L_0$ (sparsity) loss\footnote{Please refer to \cite{louizos2018learning} for $L_0$ loss and \emph{BinConcrete} function.} to encourage higher sparsity explicitly. The overall loss is a weighted sum of both: $L_{\text{\emph{total}}} = L_{\text{\emph{nll}}} + \lambda L_{\text{\emph{sparsity}}}$. The weighting coefficient $\lambda$ controls the final sparsity. When the $i$-th head is decided to be pruned ($g_i=0$), we remove the parameters corresponding to the head $\{\textbf{W}_i^Q, \textbf{W}_i^K, \textbf{W}_i^V, \textbf{P}_i^K, \textbf{P}_i^V, \textbf{W}_i^O \}$. Concurrently, their corresponding computations are removed.

% \vspace{-5pt}
% \begin{align}
%   L_{\text{\emph{sparsity}}} &= \sum_{k=i}^{L}\sum_{i=1}^{H}\text{Sigmoid}(\pi_i - \beta\log\frac{-\gamma}{\zeta}) \\
%   L &= L_{\text{\emph{nll}}} + \lambda L_{\text{\emph{sparsity}}}
% \end{align}

% \begin{figure}[t]
%   \centering
%   \vspace{5pt}
%   \includegraphics[width=1.0\linewidth]{figure/dynamics2.png}
%   \caption{Training loss at the beginning of the pruning. (A, B, C) represents whether the proposed three techniques are used. Best viewed in color. Section \ref{sec:method:pruner} explains the details.}
%   \label{fig:dynamics}
%   \vspace{-10pt}
% \end{figure}

% ======================================================================================== %

\subsection{Techniques for Head Pruning}\label{sec:method:pruner}

Pruning begins from the converged model that is previously trained without an augmented gating mechanism; therefore, the addition of attention head gating excessively changes the activation statistics and training dynamics. When this discrepancy is combined with the unique characteristics of All-att, we observe a significant performance drop and a severe instability of the pruning process, particularly in the initial training phase. To overcome the difficulties, we introduce three techniques to overcome the difficulties of pruning on All-att models.

First, we linearly increase the sparsity loss coefficient $\lambda$ from zero to the desired value. Gradual increase of $\lambda$ prevents the $L_0$ loss to overly disturb the network adapting to the stochastic activation at the beginning of the pruning process. Note that the $L_0$ objective is a powerful pressure that can be always achieved by decreasing the gating parameter values, which leads to the consistent increase of the sparsity.

Second, we initialize gating parameters $\Pi$ to a large positive value to bias the sampled stochastic gates to be opened ($g_i=1$) at the beginning. The zero initialization opens a gate with only 50\% probability. In that case, upper layers only receive abruptly reduced information and quickly loss the existing well-trained internal structure. We initialize $\Pi$ to 2, which takes about an 88\% probability of gate to be opened.

Third, as expressed in Eq.(\ref{equation:gating_projection}), we scale the output inversely proportional to the $(\text{1-sparsity})$. The scaling factor $s_g$ compensates for the masked portion and maintains the statistics after gating is applied. Recently, attention head dropout \cite{zhou2020scheduled, zhang2021stochastic} has been introduced with similar scaling, but their scaling is used for regularization during training. We found that this technique greatly stabilizes the training dynamics, especially after the training is stabilized by the above two methods. Without output scaling, we observe a consistent increase in the training loss.

% =========================================================================================== %
\section{Experimental Results}
\label{sec:experiment}
% ======================================================================================== %
\subsection{Setup}

% ======================================================================================== %
\subsubsection{Datasets and Model Architecture}

We evaluate the performance on WikiText-103 \cite{merity2016pointer} word-level LM and Text8 \cite{text8} character-level LM benchmarks. The pre-processing of datasets follows common practice \cite{dai2019transformer}. The performance is reported in perplexity (\emph{ppl}) for WikiText-103 and bit-per-character (\emph{bpc}) for Text8. Lower is better for both \emph{ppl} and \emph{bpc}.

\begin{table}[t!]
    \centering
    \def\arraystretch{1.}
    \caption{Effects of attention head pruning on the WikiText-103 test dataset. The number of parameters reported does not include token embedding and output projection weight. The embedding and projection occupy 19.6M parameters.}
    \vspace{10pt}
    \resizebox{\linewidth}{!}{
    \begin{tabular}{c|ccc|c}
        \toprule
        $\lambda$  & Sparsity (\%) & \#Params(w/o emb.) && \emph{ppl}\\
        \midrule
        0(base)  & 0 & 54.6M && 23.24 \\
        0.01  & 17.2 & 45.2M && 23.45\\
        0.015  & 32.8 & 36.7M && 24.07\\
        0.02  & 43.8 & 30.7M && 24.85\\
        % 0.025 & 25.30 && 50.8 & 26.9M \\
        \bottomrule
    \end{tabular}
    }
    \label{tab:performance_wt103}
    \vspace{-5pt}
\end{table}

\begin{table}[t]
    \centering
    \def\arraystretch{1.0}
    \caption{Effects of attention head pruning on the Text8 test set. The embedding and projection occupy 0.6M parameters.}
    \vspace{10pt}
    \resizebox{\linewidth}{!}{
    \begin{tabular}{c|ccc|c}
        \toprule
        $\lambda$ & Sparsity (\%) & \#Params(w/o emb.) && \emph{bpc} \\
        \midrule
        0(base) & 0 & 40.9M && 1.199\\
        0.01 & 15.6 & 34.5M && 1.204\\
        0.015 & 27.1 & 29.8M && 1.218\\
        0.02 & 37.5 & 25.6M && 1.234\\
        \bottomrule
    \end{tabular}
    }
    \label{tab:performance_t8}
    \vspace{-10pt}
\end{table}

% \hline\hline
% 0 & 3.161(23.59) & (7/8 heads) & 47.7M \\
% 0 & 3.176(23.95) & (6/8 heads) & 40.9M\\
% 0 & 3.197(24.46) & (5/8 heads) & 34.1M\\
% 0 & 3.227(25.21) & (4/8 heads) & 27.3M\\

\begin{table}[t]
    \centering
    \caption{Ablation study of each proposed technique. The results are evaluated on the WikiText-103 test dataset. $\lambda$ is set to 0.02. "vanilla" means no technique is applied.}
    \vspace{10pt}
    % \resizebox{\linewidth}{!}{
    \begin{tabular}{ll|c}
        \toprule
         Ablation & \phantom{..................} &   $\Delta$\emph{ppl} \\
        \midrule
         - $\lambda$ warm-up &(X,O,O)&   +0.12 \\
         - proper gate initialization &(O,X,O)&   +0.61 \\
         - attention head output scaling &(O,O,X)&   +1.25 \\
         - all (vanilla) &(X,X,X)&   +1.48 \\
        \bottomrule
    \end{tabular}
    % }
    \label{tab:ablation}
    \vspace{-10pt}
\end{table}

% Tran.XL\cite{dai2019transformer} & 3.179(24.03) & 8 heads & 41.1M \\
% (standard) & & dim 410 & 151.1M \\
% Tran.XL\cite{nvidia-txl} & 3.146(23.24) & 8 heads & 54.6M \\
% (standard) & & dim 512 & 192.0M \\
% All-att\cite{sukhbaatar2019augmenting} & 3.02(20.6) & 8 heads & 113.3M \\
% (w/adap.span) & & dim 512 & 132.5M \\

% xl width 480 head 8 -> 48.0M / 176.8M / 24.90
% xl width 448 head 8 -> 41.8M / 162.1M / 25.16
% xl width 416 head 8 -> 36.1M / 147.7M / 25.57
% xl width 384 head 8 -> 30.7M / 133.8M / 26.32
% xl width 352 head 8 -> 25.8M / 120.3M / 26.70
% xl width 320 head 8 -> 21.3M / 107.3M / 27.68

The baseline model is a variant of All-attention Transformer. The model adopts a pre-norm instead of a post-norm and omits the adaptive-span \cite{sukhbaatar2019adaptive} mechanism. The configuration of the transformer layer is as follows: the hidden dimension $d=512$, number of heads $H=8$, and number of persistent vectors $N=2048$. We stack 16 layers for WikiText-103 and 12 layers for Text8.

% \begin{table*}[t]
%   \centering
%   \caption{Number of pruned attention heads per layer. Overall, 16 layers $\times$ 8 = 128 heads and 12 $\times$ 8 = 96 heads exist in models trained on WikiText-103 (upper) and Text8 (lower), respectively. A layer with a smaller index is closer to the input.}
%   \vspace{-5pt}
%   \def\arraystretch{1.0}
%   \begin{tabular}{c|cccc cccc cccc cccc|c|c}
%     \toprule
%     % \diagbox{$\lambda$}{layer}& 1 & 2 & 3 & 4 & 5 & 6 & 7 & 8 & 9 & 10 & 11 & 12 & 13 & 14 & 15 & 16 & Total & Sparsity (\%) \\
%     $\lambda$ \textbackslash layer & 1 & 2 & 3 & 4 & 5 & 6 & 7 & 8 & 9 & 10 & 11 & 12 & 13 & 14 & 15 & 16 & Total & Sparsity (\%) \\
%     \midrule
%     0.01  & 0 & 5 & 1 & 2 & 2 & 2 & 2 & 1 & 2 & 1 & 0 & 0 & 1 & 0 & 1 & 2 & 22 & 17.2 \\
%     0.02  & 2 & 5 & 3 & 3 & 5 & 3 & 3 & 4 & 3 & 4 & 3 & 3 & 4 & 4 & 3 & 4 & 56 & 43.8 \\
%     \hline
%     0.01  & 0 & 3 & 2 & 2 & 1 & 3 & 2 & 2 & 0 & 0 & 0 & 0 &  &  &  &  & 15 & 15.6 \\
%     0.02  & 2 & 4 & 3 & 4 & 4 & 4 & 3 & 4 & 3 & 2 & 2 & 1 &  &  &  &  & 36 & 37.5 \\
%     \bottomrule
%     \end{tabular}
%     \vspace{-10pt}
%   \label{tab:num_heads}
% \end{table*}

% ======================================================================================== %
\subsubsection{Training Details}

We first train the baseline model with full attention heads. We utilize the LAMB optimizer with a batch size of 96 for WikiText-103 and 64 for Text8. The sequence length $T$ and memory length $S$ are both set to 192 for WikiText-103 and 512 for Text8. We apply linear warm-up on learning rate for 4K iterations. The learning rate increases to $1\text{e-}2$ and gradually decreased by cosine learning rate scheduling to $1\text{e-}4$. The training requires 160K iterations to converge. We use a dropout rate of 0.2 for attention matrices, 0.1 for embedding and hidden activation.

We start pruning from the converged baseline. Pruning follows identical training configurations except for the learning rate, which increases to $1\text{e-}3$ and gradually decreased to $1\text{e-}4$. The pruning requires additional 80K iterations of training. As explained in Sec.\ref{sec:method:pruner}, we warm-up $\lambda$ from zero to the desired value for the first 4K iterations. After 16K iterations, we stop training the gating parameters, so that the training continues without randomness for the remaining steps. Without this fixation, we observe that the sparsity continues to increase because of the influence of $L_0$ loss becomes too large, which causes the network much difficult to be fine-tuned. We explore $\lambda=\{1.0, 1.5, 2.0\}\cdot 10^{-2}$ to control the trade-off between the sparsity and performance.

\begin{figure}[!t]
  \centering
  \includegraphics[width=\linewidth]{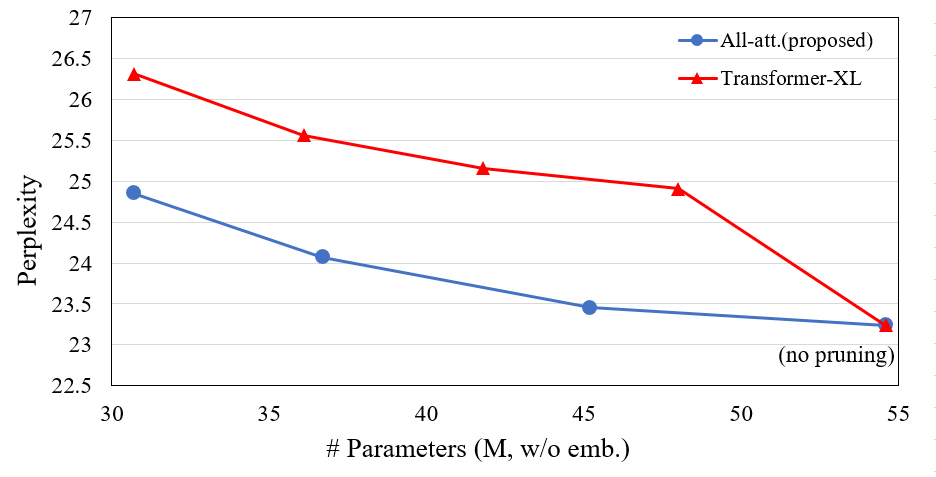}
  \caption{Perplexity of the pruned All-attention Transformer and TXL. The former shows a better parameter efficiency.}
  \label{fig:result}
  \vspace{-10pt}
\end{figure}

% ======================================================================================== %
\subsection{Results on Language Modeling}

Tables \ref{tab:performance_wt103} and \ref{tab:performance_t8} show the results of attention head pruning on two benchmarks. As expected, the number of parameters linearly decreases as sparsity increases on All-att models. We observe a clear trade-off between sparsity and performance for both datasets.

To compare with the original Transformer architecture, we train TXL models with reduced dimensions under the same configuration. Each TXL model utilizes the same number of layers and heads, whereas the hidden dimension decreases from 512 by 32 in sequence. Both All-att and TXL baselines achieve almost same perplexity and the parameter size. Figure \ref{fig:result} shows that All-att models with attention head pruning achieve substantially better parameter efficiency than the TXL models. For  example, pruned  All-att model with 43\% sparsity (30.7M) achieves similar perplexity as TXL with only 25\% sparsity (47.9M).

We empirically show that the proposed three methods each contribute to the improvement. Table \ref{tab:ablation} compares the effect of each technique by ablation. The most influential change is achieved by output scaling (+1.25), however, the other two also take a portion of the improvement. All-att model without proposed techniques (denoted as "vanilla"), is expected to suffer from a similar level of performance degradation as TXL, which implies that the potential of pruning efficiency on All-att cannot be fully utilized without our techniques.

% =========================================================================================== %
\section{Conclusion}
\label{sec:conclusion}
In this paper, we introduced layer-wise attention head pruning for All-attention Transformer models and proposed three techniques to reduce the performance degradation of the pruned model and stabilize the pruning process. Experiments on language modeling demonstrate that the proposed method achieves a better performance than traditional Transformer models with a comparable number of parameters.

% =========================================================================================== %
\vfill\pagebreak

\bibliographystyle{IEEEbib}
\bibliography{references}

\end{document}